\documentclass[runningheads]{llncs}
\usepackage{eccv}

\usepackage{eccvabbrv}
\usepackage[table,xcdraw]{xcolor}
\definecolor{lightgreen}{rgb}{0.88, 1, 0.88}
\usepackage{graphicx}
\usepackage{booktabs}
\usepackage{amsmath}
\usepackage{amssymb}
\usepackage{subcaption}

\usepackage[accsupp]{axessibility}  % Improves PDF readability for those with disabilities.

\usepackage{hyperref}

\usepackage{orcidlink}

\begin{document}

\title{Following the Flow: Advection-Consistent Modeling for Event-based Small Object Detection}

\titlerunning{Advection Consistency for Weak Event Responses}

\author{
Wen Guo\inst{1}\orcidlink{0000-0002-3691-4942}
\and
Fulong Cai\inst{1}\orcidlink{0009-0001-5383-4354}
\and
Wuzhou Quan\inst{2}\orcidlink{0000-0002-5593-2054}\thanks{Corresponding author.}
}
\authorrunning{W. Guo et al.}
\institute{
School of Information and Electronic Engineering,
Shandong Technology and Business University,
Yantai 264005, China\\
\email{\{wguo,2024410008\}@sdtbu.edu.cn}
\and
Nanjing University of Aeronautics and Astronautics,
Nanjing 210016, China\\
\email{q.wuzhou@gmail.com}
}

\maketitle

\begin{abstract}
Event cameras enable high-frequency visual perception with microsecond latency, offering advantages for dynamic scenes. However, event-based small object detection remains challenging due to sparse asynchronous measurements and weak object responses that are easily disrupted by noise. Limited spatial support causes small-object signals to lose temporal continuity, resulting in fragmented and unstable predictions.
To address this issue, we propose a physics-guided advection-consistent modeling framework, termed PACT, which formulates event evolution as a motion-driven feature transport process. Instead of relying solely on local spatio-temporal aggregation, PACT propagates features along estimated velocity fields and enforces trajectory-level consistency through advection constraints. This design preserves weak event responses over time and prevents their degradation under complex background interference.
Technically, PACT integrates motion-aware feature extraction with a differentiable advection-based transport operator, enabling coherent motion representation and effective noise suppression during temporal evolution.
Extensive experiments on benchmark event-based datasets demonstrate that PACT consistently outperforms state-of-the-art methods, achieving improvements of 20.72\% in IoU and 15.03\% in accuracy while maintaining comparable computational efficiency. 
The code is publicly available at \url{https://github.com/fulongcai/PACT}.
\end{abstract}
\section{Introduction}
\label{sec:intro}

% 事件相机是一类新一代视觉传感器，它们以异步方式响应场景亮度的变化，输出具有微秒级时间分辨率与宽动态范围的测量。这些特性使其特别适合捕捉传统帧式相机难以观测的快速动态过程，以及由变化驱动的细粒度时间变化。不同于由图像序列构成的数据，事件流可以被理解为一种时空信号，其结构会随时间持续演化。尽管这种表征提供了很高的时间精度，但也使成像呈现稀疏、不规则且对噪声敏感的特点。因此，由微小或弱对比实体所诱发的事件响应即便遵循一致的运动学模式，也常常在空间上呈碎片化、在时间上呈间歇性，尤其在强噪声背景活动存在时更为明显。能否准确建模并维持这些微弱事件响应，因而成为在复杂噪声条件下实现可靠且具泛化能力的事件感知的关键。

% 事件相机是什么，怎么生成数据的
Event cameras represent a new generation of vision sensors that respond asynchronously to changes in scene brightness, producing measurements with microsecond level temporal resolution and a wide dynamic range \cite{lichtsteiner2008temporal,gallego2022survey}.
These properties make event cameras particularly effective at capturing rapid scene dynamics and fine-grained temporal variations driven by brightness changes, which are difficult to observe with traditional frame-based cameras \cite{kim2016realtime,pan2022high}.
% 事件相机生成的数据和其他范式的数据有什么区别
Instead of being a sequence of images, an event stream can be interpreted as a spatio-temporal signal whose structure continuously evolves over time.
While this representation provides high temporal resolution, it also makes the resulting measurements sparse, irregular, and sensitive to noise \cite{shariff2024event}.
% 事件相机生成的数据有什么缺点，引出我们要解决的问题，就是微小实体形成的弱小响应在事件流中难以保持
Thus, event responses triggered by small or low-contrast entities are often fragmented in space and intermittent in time, even when their motion follows coherent kinematic patterns, especially under strong background activity and noise \cite{mondal2021moving}.
% 因此，能否准确建模这些微小实体的真实响应，决定了事件感知在强噪声背景下的可靠性与泛化能力。
Accurately modeling and maintaining these weak event responses is therefore critical for reliable and generalizable event-based perception under challenging conditions with strong noise.

% 现有的事件表征与建模方法总体上主要遵循两类范式。一类方法将异步事件聚合为类似帧的表征，使后续处理可以沿用传统的图像模型。尽管这种策略简化了实现流程，但它不可避免地将事件随时间连续演化的过程压缩为静态模式，恰恰抹去了区分微弱事件响应与背景噪声所必需的时间线索。另一类方法依赖脉冲神经网络来保持时间连续性，并通过神经元的反复激活来实现证据累积。然而，微弱事件响应通常稀疏、间歇，并且伴随快速的空间漂移，这使其难以触发稳定的重复激活。结果是，这些微弱响应在统计上会逐渐与随机噪声触发趋于一致，导致判别结构随时间衰减甚至消失。两类范式共同的局限在于，它们对时间连续性的建模多为隐式方式，并未显式约束事件响应如何随时间传播。对于在任意单个时刻都呈碎片化的微弱实体，仅仅保留时间信息并不足够。缺乏显式的传播模型时，无论时间模型容量多强，微弱轨迹都难以在强背景活动下维持为跨时间一致的结构。
Existing approaches generally fall into two paradigms.
Frame-based pipelines first aggregate events into static representations \cite{Maqueda2018event, mitrokhin2020learning,yang2022querydet}. 
This aggregation inevitably collapses the continuous temporal evolution and removes cues needed to distinguish weak signals from noise.
Alternatively, Spiking Neural Networks (SNNs) \cite{neftci2019surrogate,su2023deep,luo2024integer,cordone2022object} perform temporal integration through neuronal dynamics, yet weak and intermittent responses often fail to elicit stable spiking activity and become indistinguishable from background noise.
A critical shared limitation is that both paradigms model temporal continuity implicitly. 
Without an explicit propagation rule, these approaches struggle to preserve fragmented weak trajectories as coherent structures under strong background activity.

% 关键在于，尽管微弱实体只会诱发稀疏且间歇的事件响应，物理实体本身仍然遵循运动连续性。这一性质从根本上将其与缺乏持久运动来源的背景噪声区分开来。在足够短的时间间隔内，同一实体诱发的事件响应往往具有因果关联，并沿着一致的运动方向持续位移，而由噪声诱发的触发则不具备这种方向上的持续性。在这一尺度下，运动连续性成为区分由实体诱发的微弱响应与随机背景活动的唯一剩余线索。 
Crucially, although weak entities induce sparse and intermittent event responses, the underlying physical motion remains continuous.
This property fundamentally distinguishes such weak responses from background noise, which lacks a persistent motion source \cite{gallego2022survey}.
Within sufficiently short time intervals, event responses induced by the same entity tend to be temporally coherent and continuously displaced along a consistent motion direction, whereas noise-induced triggers exhibit no such directional persistence.
In this regime, motion continuity becomes the only remaining cue capable of separating entity-induced weak responses from random background activity.

% 为形式化这一差别，我们寻求一种尽可能弱的物理约束，它能够刻画运动连续性，同时不对目标外观、形状或强度施加更强假设。在短时间尺度上，事件响应的演化可以被很好地近似为在局部速度场驱动下的输运过程，而形变或加速度等高阶效应可以忽略。这种由速度场驱动的时空演化与物理学中的平流输运直接对应。更重要的是，平流提供了一条最小且不可回避的约束：源自同一运动实体的响应在一致的输运场作用下会保持对齐，而在缺乏一致运动来源时，零散的噪声触发无法满足这种一致性。
To formalize this distinction, we adopt the weakest physical constraint that captures motion continuity without imposing stronger assumptions on appearance, shape, or intensity.
Over short time scales, the evolution of event responses can be well approximated by advection under a local velocity field \cite{gallego2018unifying,benosman2013event,zhu2019unsupervised}, while higher-order effects such as deformation or acceleration are negligible.
This velocity-driven spatiotemporal evolution naturally aligns with advection in physics.
Importantly, advection imposes a minimal and unavoidable constraint: responses originating from the same moving entity remain under a consistent transport field, whereas scattered noise triggers, lacking a coherent motion source, cannot satisfy such consistency.

% 基于这一观察，我们引入平流一致性约束，用以显式建模事件响应随时间的传播。在该约束下，能够沿局部速度场被稳定输运的响应会在时间上保持对齐并相互强化，而违反平流一致性的响应则会自然失去对齐并逐步被抑制。因此，平流一致性为在高噪声事件流中保留微弱但具有物理意义的轨迹提供了一种可操作的判据。
Motivated by this observation, we introduce an advection-consistency constraint to explicitly model the propagation of event responses over time.
Under this constraint, responses that can be consistently transported along a local velocity field remain temporally aligned and accumulate coherently, whereas responses that violate advection consistency lose alignment and are progressively suppressed.
Advection consistency therefore provides an operational criterion for preserving weak yet physically meaningful trajectories in highly noisy event streams.

% 在上述形式化基础上，我们提出了 PACT，一种以物理一致的平流一致性输运机制为核心的框架，用于维持微弱且运动一致的事件结构。PACT 由三个协同组件构成。轨迹引导特征提取模块 T FE 从稀疏事件特征中提炼感知运动的表征，用于支撑轨迹层面的推理。基于平流的轨迹一致性模块 ATC 估计局部一致的速度场，通过轨迹一致的速度损失进行监督，并借助平流一致性验证时间一致性。最后，平流一致性特征重构模块 A FR 沿估计的速度场对特征进行对齐与重建，将碎片化响应恢复为连续的轨迹结构，而在缺乏一致输运的情况下，背景杂波会逐渐衰减。我们在事件小目标检测基准上对 PACT 进行了评估，将其作为上述表述的一种具体实例，结果表明该框架在挑战性条件下能够恢复微弱轨迹并抑制噪声。尽管实验在检测任务背景下开展，但所提出的基于平流的表述并不依赖于任何特定任务，它为在事件流中维持微弱时空结构提供了一种通用机制。
Building upon this formulation, we propose \textbf{PACT}, a framework that leverages a \underline{P}hysics-consistent \underline{A}dvection-\underline{C}onsistent \underline{T}ransport mechanism to maintain weak yet motion-consistent event structures.
PACT consists of three complementary components that work in concert.
The Trajectory-Guided Feature Extraction (T-FE) module distills motion-aware features from sparse event measurements to support trajectory reasoning.
The Advection-based Trajectory Consistency (ATC) module estimates a locally consistent velocity field, supervised by a trajectory-consistent velocity loss, and enforces temporal coherence via advection consistency.
Finally, the Advection-Consistent Feature Reconstruction (A-FR) module aligns and reconstructs features along the estimated velocity field, turning fragmented responses into continuous trajectory structures, while background clutter is progressively suppressed when transport consistency is absent.
In practice, weak event responses are most prominent in small-object scenarios, where sparse and intermittent activations are easily dominated by background activity.
We evaluate PACT on event-based small object detection benchmarks as a concrete instantiation of the proposed formulation, demonstrating its ability to recover weak trajectories and suppress noise under challenging conditions.
Although our experiments are conducted in the context of detection, the proposed advection-based formulation is defined independently of any specific task and provides a general mechanism for maintaining weak spatio-temporal structures in event streams.

\section{Related Work}

\subsection{Event Representation and Temporal Association}
A common way to process event streams is to convert asynchronous events into regular tensors so that standard backbones can be applied \cite{Maqueda2018event}.
Early designs aggregate polarities or counts within a window, while later ones inject timing cues, such as time surfaces and kernelized timestamp encoding \cite{lagorce2016hots, gehrig2019end}.
Reconstruction-based pipelines further map events to intensity-like proxies to reintroduce photometric cues \cite{rebecq2021high}.
To preserve temporal structure more explicitly, voxel-based methods discretize events into spatiotemporal volumes and apply 3D operators or recurrent updates to extend the temporal support \cite{zhu2019unsupervised, perot2020learning, gehrig2023recurrent}.
Other approaches keep state evolution with spiking dynamics or sparse relational reasoning \cite{su2023deep, zhang2022spiking, schaefer2022aegnn, bi2019graph}.
Despite different forms, most methods rely on implicit continuity.
Cross-temporal linkage is largely driven by local neighborhoods or short-range propagation.
Under strong background activity, weak responses drift and appear intermittently, so evidence fragments rather than accumulating into motion-consistent traces.

\subsection{Physics-guided Dynamics}
To stabilize motion cues in low-SNR event streams, many works introduce physical priors.
A classic line is contrast maximization, which warps events to a reference time under a motion model and optimizes the motion parameters to sharpen the compensated accumulation \cite{gallego2018unifying, stoffregen2019event, mitrokhin2018event}.
Other formulations propagate sparse evidence with dynamical processes, including diffusion-like smoothing and probabilistic motion modeling \cite{wang2025object, sekikawa2023live}.
Diffusion can denoise, but isotropic smoothing may also spread clutter and weaken directional patterns.
Advection provides a motion-aligned alternative.
A local velocity field specifies how information should move over short time spans \cite{gallego2018unifying, benosman2013event, zhu2019unsupervised}.
PACT builds on this view and models representation evolution as advection-consistent transport, so aligned responses persist and accumulate, while inconsistent triggers lose alignment and are suppressed.

\section{Methods}
\label{sec:method}

\subsection{Advection-Consistent Representation}
\label{sec:adv_rep}
\begin{figure*}[t]
  \centering
  \includegraphics[width=\textwidth]{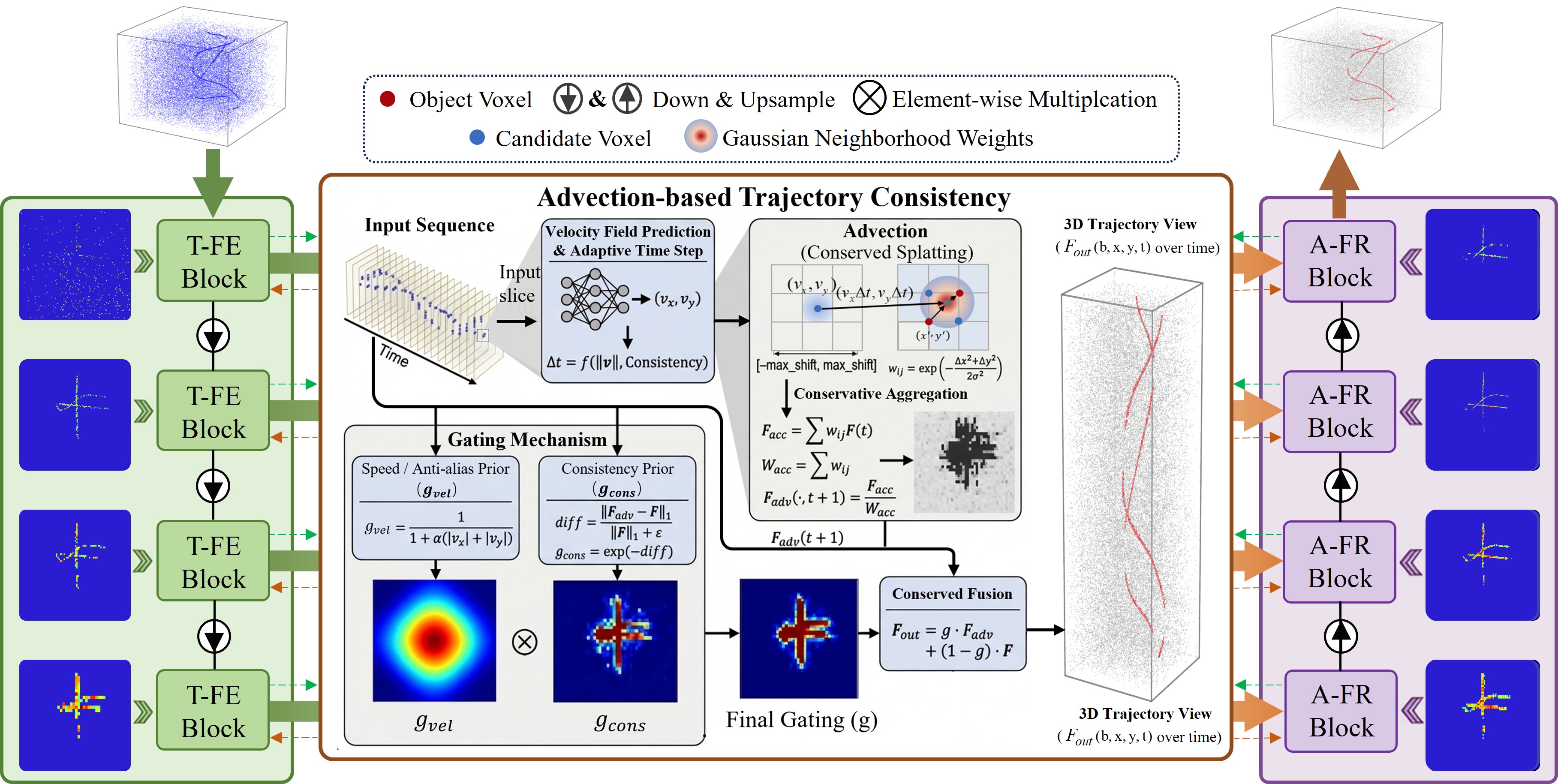}
  \caption{Overall architecture of PACT. 
  The encoder stacks four T-FE modules.
ATC estimates velocity fields and consistency scores at each scale.
The decoder uses four A-FR modules with skip fusion to reconstruct temporally continuous trajectories.}
  \label{fig:overall}
\end{figure*}
Given sparse, asynchronous event streams, informative responses often appear as sparse activations with fragmented temporal support, yet reveal coherent structure when traced over time.
Coherent activations remain mutually alignable when propagated by a local motion field, whereas spurious triggers from background activity quickly lose temporal consistency over time.
This motivates an advection-consistent representation, where the feature field is approximately conserved under advection driven by the local motion field.

We treat the sparse voxel features as samples of a $C$-channel field $u(\mathbf{x},t)$ defined over space and time. 
Within a short time interval, its evolution is approximated by the advection equation~\cite{benosman2013event}: 
\begin{equation}
\frac{\partial u(\mathbf{x},t)}{\partial t}+\mathbf{v}(\mathbf{x},t)\cdot\nabla u(\mathbf{x},t)\approx0,
\end{equation}
where $\mathbf{v}(\mathbf{x},t)$ denotes a local velocity field. 
Along a characteristic curve $\mathbf{x}(t)$ satisfying $\frac{d\mathbf{x}(t)}{dt}=\mathbf{v}(\mathbf{x}(t),t)$, the transported feature is approximately preserved,
\begin{equation}
\frac{d}{dt}u(\mathbf{x}(t),t)\approx 0.
\end{equation}
For a discrete temporal step $\tau$ and locally constant velocity, the transport admits a displacement operator: 
\begin{equation}
\mathcal{T}_{\mathbf{v}}(u)(\mathbf{x},t)\;=\;u(\mathbf{x}-\tau\mathbf{v},t-\tau).
\end{equation}
Advection consistency corresponds to a small deviation under $\mathcal{T}_{\mathbf{v}}$~\cite{zhu2019unsupervised}. 
We therefore measure the transport residual as:
\begin{equation}
\mathcal{R}_{adv}(\mathbf{x},t)\;=\;\left\|u(\mathbf{x},t)-\mathcal{T}_{\mathbf{v}}(u)(\mathbf{x},t)\right\|,
\end{equation}
where $\|\cdot\|$ is a channel-wise discrepancy measure.
Throughout the pipeline, $\mathcal{R}_{adv}$ provides a consistency signal. 
Features with smaller residuals receive higher confidence and are propagated more reliably under the estimated local velocity field.

\subsection{Trajectory-Constrained Feature Encoding}
\label{sec:traj_encoding}

\begin{figure*}[t]
  \centering
  \begin{subfigure}{0.42\textwidth}
    \centering
    \includegraphics[width=\linewidth]{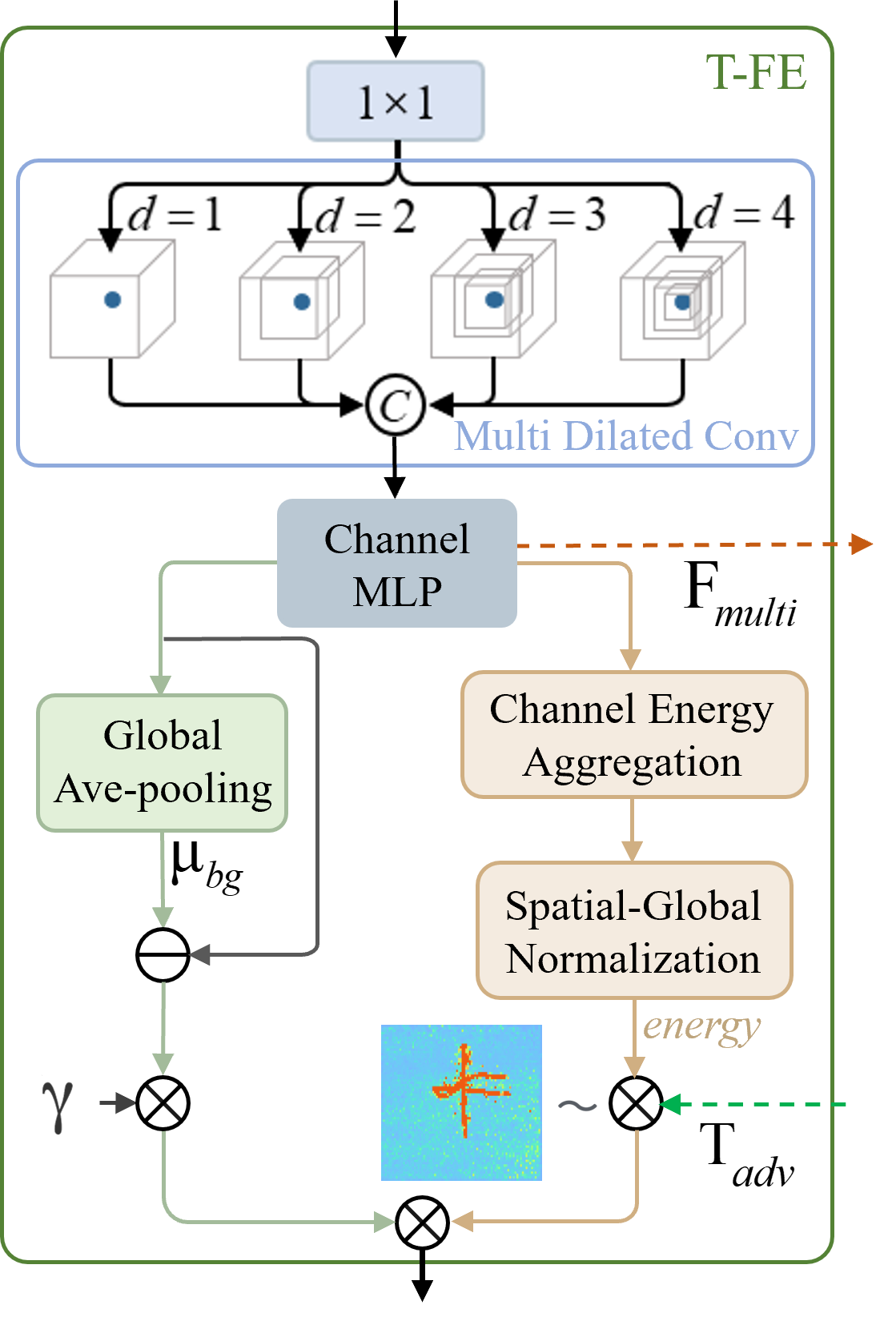}
    \caption{T-FE module.}
    \label{fig:tfe}
  \end{subfigure}
  \hfill
  \begin{subfigure}{0.55\textwidth}
    \centering
    \includegraphics[width=\linewidth]{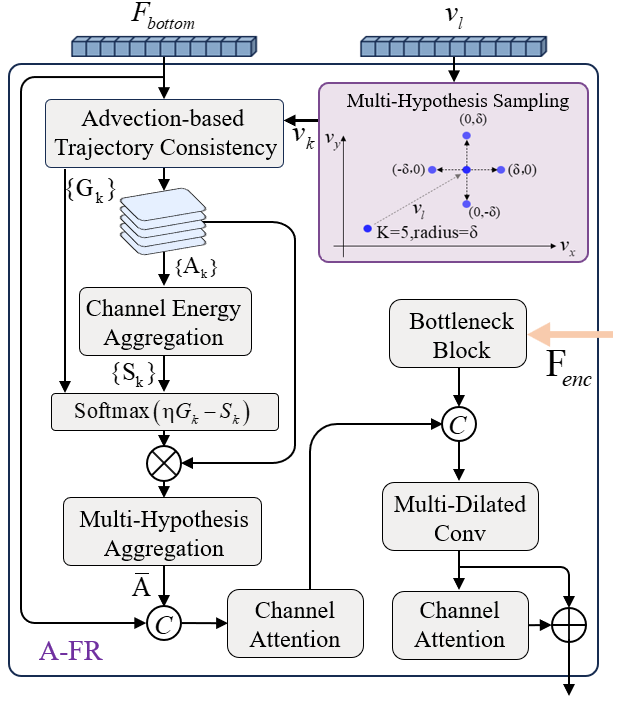}
    \caption{A-FR module.}
    \label{fig:afr}
  \end{subfigure}
  \caption{Architectures of the proposed modules. (a) T-FE filters and enhances sparse responses under trajectory-consistency guidance. (b) A-FR propagates features along the estimated velocity field to reconstruct continuous trajectories while suppressing incoherent triggers.}
  \label{fig:modules}
\end{figure*}

As illustrated in Fig.~\ref{fig:overall}, the pipeline follows an encoder--decoder architecture with multi-scale sparse processing.
The input is a sparse voxel tensor $\mathrm{X}=\{\mathrm{C},\mathrm{A}\}$, where $\mathrm{C}$ stores voxel indices $(b,x,y,t)$ and $\mathrm{A}$ stores normalized attributes $(\tilde{x},\tilde{y},\tilde{t},p)$ with polarity $p$.
The encoder progressively extracts motion-aware features using four stages of sparse processing, producing a pyramid $\{\mathrm{F}^{i}_{enc}\}_{i=1}^{4}$.
Rather than treating motion priors as an external post-processing step, we impose trajectory consistency \emph{inside} feature encoding.
At each scale, the representation is shaped to preserve components that remain coherent under transport, so that the subsequent decoding can propagate and connect weak responses more reliably.

At each scale, as shown in Fig.~\ref{fig:modules}(a), T-FE aggregates spatiotemporal context to support transport assessment.
Specifically, we form $\mathrm{F}_{multi}$ via multi-branch sparse convolutions with dilation rates $\mathcal{D}=\{1,2,3,4\}$.
ATC then evaluates transport feasibility on $\mathrm{F}_{multi}$ and outputs participation weights for subsequent propagation.

At the core is a trajectory-consistency evaluator that predicts a local displacement field and measures the agreement between the observed feature and its transported prediction.
For each active voxel $i$ with feature $\mathrm{F}_i$, we predict a bounded 2D displacement $\Delta\mathbf{p}_i$ on the spatial grid:
\begin{equation}
\Delta\mathbf{p}_{i}=\mathrm{v}_{max}\tanh(\mathrm{MLP}(\mathrm{F}_{i})), \quad \mathbf{p}_{i}^\prime=\mathbf{p}_{i}+\Delta\mathbf{p}_{i},
\end{equation}
where $\mathrm{v}_{max}$ limits the maximum displacement for stability.

Since $\mathbf{p}'_{i}$ is generally off-grid, we obtain $\mathrm{F}_{adv}$ by Gaussian-kernel sampling on the $(x,y)$ lattice with normalized weights.

We quantify trajectory consistency using a gating weight $g$ that combines the agreement between $\mathrm{F}_{adv}$ and $\mathrm{F}$ and a mild penalty on large velocities,
\begin{equation}
g
=\exp\left(-\frac{\|\mathrm{F}_{adv}-\mathrm{F}\|_{1}}{\|\mathrm{F}\|_{1}+\varepsilon}\right)\odot
\frac{1}{1+\alpha\left(|v_{x}|+|v_{y}|\right)}.
\end{equation}
This weight yields a conservative fusion between the transported prediction and the original observation,
\begin{equation}
\mathrm{T}_{adv}=g\cdot \mathrm{F}_{adv}+(1-g)\cdot \mathrm{F}.
\end{equation}
Intuitively, $g$ acts as a trajectory-consistency score: components that remain stable under transport receive stronger support from the transported prediction, while inconsistent components are prevented from dominating the representation.
We use $\mathrm{T}_{adv}$ and $g$ to constrain encoding at each scale, producing features that are better aligned with the estimated velocity field and thus more amenable to transport-based propagation.

\subsection{Advection-Guided Trajectory Propagation}
\label{sec:adv_propagation}

% \begin{figure*}[hbt!] 
%   \centering
%   \includegraphics[width=0.45\linewidth]{images/A-FR block.png}
%   \caption{Architecture of the proposed A-FR module. The A-FR module propagates motion semantics along the local velocity field to reconstruct continuous trajectories, while suppressing spurious responses induced by random triggers.}
%   \label{fig:A-FR}
% \end{figure*}
The decoder aims to transform deep motion-aware features into temporally continuous trajectory structures. 
This is challenging in event data because coherent activations are intermittent, and naive upsampling tends to amplify isolated triggers.
We therefore perform decoding as \emph{advection-guided propagation} by extending features along the velocity field to connect fragmented responses over time, while naturally down-weighting components that do not admit coherent transport.

At each decoding stage, as illustrated in Fig.~\ref{fig:modules} (b), we take the lower-resolution feature $\mathrm{F}_{bottom}$ and the corresponding velocity guidance $\mathrm{v}_{i}$ from the encoder, and perform transport-based alignment.
Because velocity estimation can be uncertain in sparsely activated regions, we form a set of perturbed hypotheses around $\mathrm{v}_{i}$,
\begin{equation}
\mathrm{v}^{k}_{i}=\mathrm{v}_{i}+\delta_{k},\quad k=1,\ldots,K,
\end{equation}
where $\delta_{k}$ are fixed small perturbations.
For each hypothesis, we compute a transported candidate and its trajectory consistency,
\begin{equation}
\{\mathrm{A}_{k},\mathrm{G}_{k}\}=\mathrm{ATC}(\mathrm{F}_{bottom},\mathrm{v}^{k}_{i}),
\end{equation}
where $\mathrm{A}_{k}$ is the transported feature and $\mathrm{G}_{k}$ is the corresponding confidence derived from transport agreement.
To avoid linking trajectories through abnormally high activations, we introduce a magnitude penalty $S_k$ and compute consistency-aware aggregation weights:
\begin{equation}
\pi_{k}=\mathrm{Softmax}\left(\eta\,\mathrm{G}_{k}-\mathrm{S}_{k}\right),
\qquad
\bar{\mathrm{A}}=\sum_{k=1}^{K}\pi_{k}\mathrm{A}_{k}.
\end{equation}
Here $S_k$ is the channel-averaged energy of $\mathrm{A}_k$.

We then combine the aligned feature with the original bottom feature to avoid information loss,
\begin{equation}
\mathrm{F}_{align}=\mathrm{CA}\Big(\mathrm{Concat}(\mathrm{F}_{bottom},\bar{\mathrm{A}})\Big),
\end{equation}
where $\mathrm{CA}$ is a lightweight channel attention.
Finally, we inject the corresponding encoder skip feature $\mathrm{F}_{enc}$ to recover spatial details. 
The skip feature is first compressed by a bottleneck transform, concatenated with $\mathrm{F}_{align}$, and refined by multi-dilated sparse convolutions with a residual connection,
\begin{equation}
\mathrm{F}_{traj}
=\mathrm{CA}\Big(
\mathrm{DConv}_{d}\big(\mathrm{Concat}(\mathrm{F}_{align},\mathrm{Bottleneck}(\mathrm{F}_{enc}))\big)
+\mathrm{F}_{align}
\Big).
\end{equation}
Repeating this process across scales produces a trajectory feature pyramid $\{\mathrm{F}^{i}_{traj}\}_{i=1}^{4}$, whose responses are progressively connected along time by advection-guided transport, yielding temporally coherent trajectory representations for downstream prediction.

\subsection{Learning Objectives}
\label{sec:objectives}
We train the network with a joint objective that combines voxel-level segmentation supervision and a regularizer on the predicted velocity field,
\begin{equation}
\mathcal{L}=(1-\lambda)\mathcal{L}_{seg}+\lambda\mathcal{L}_{vel},
\end{equation}
where $\lambda$ balances the two terms.
We use binary cross-entropy for $\mathcal{L}_{seg}$ over predicted mask probabilities.

Dense flow annotations are unavailable in typical event settings.
We therefore construct \textbf{pseudo velocity targets} via sparse temporal matching between foreground voxels across nearby time slices.
For a foreground voxel $i$ at time $t$ with center $\mathrm{p}_{i}$, we find a nearest matched voxel $j$ from the same instance at a neighboring time $t'$, and define:
\begin{equation}
\mathrm{v}^{*}_{i}=\frac{\mathrm{p}_{j}-\mathrm{p}_{i}}{t'-t}.
\end{equation}
We supervise the predicted velocity with the Smooth-$\ell_{1}$ loss,
\begin{equation}
\mathcal{L}_{vel}=
\frac{1}{|\mathcal{V}|}\sum_{i\in\mathcal{V}}
\mathrm{SmoothL1}\left(\mathbf{v}_{i}-\mathbf{v}^{*}_{i}\right),
\end{equation}
where $\mathcal{V}$ is the set of supervised foreground voxels.
This regularizer stabilizes the learned transport field and improves temporal coherence, while the main supervision remains on the final task output.

\section{Experiments}
\subsection{Implementation Details}

All experiments are conducted on the EV-UAV dataset~\cite{chen2025event}, which contains 147 sequences and over 2.3M event-level annotations.
The targets are extremely small, with an average size of $6.8\times5.4$ pixels, and appear under complex backgrounds and challenging illumination.
We follow the official split with 99 training and 24 test sequences.
All compared methods are retrained on the same split.
For fairness, all methods are evaluated on the same raw event streams, annotations, and metrics, while each baseline keeps its native event representation.

Input streams are sliced into 8s windows and voxelized at a temporal resolution of 1 ms.
We train PACT for 50 epochs using Adam~\cite{kingma2014adam} with an initial learning rate of $10^{-3}$ and decay on an NVIDIA RTX 4060 Ti GPU.
For all experiments, we use the same hyperparameter setting: $\alpha=0.25$, $\epsilon=10^{-6}$, $K=5$, $\delta_k=0.6$, $\eta=0.75$, and $\lambda=0.3$.
We report IoU and Acc for event-level segmentation quality, and $P_d$ and $F_a$ for localization performance.

\subsection{Benchmark Comparisons}
\begin{table*}[t]
    \centering
    \caption{Quantitative comparison of the proposed method to state-of-the-art methods.
    The \textbf{bold} and the \underline{underline} represent the best and second-best performance, respectively.}
    \label{tab:comparison}
    
    \resizebox{\textwidth}{!}{%
        \begin{tabular}{llcccccccc}
            \toprule
            Methods & Publications & Event Rep. & Temporal & $IoU(\%)\uparrow$ & $ACC(\%)\uparrow$ & $P_d(\%)\uparrow$ & $F_a(10^{-4})\downarrow$ & \#Params. & Runtime(ms) \\ 
            \midrule
            
            SSD & ECCV 2016 & Event Count & No & 25.31 & 28.56 & 26.31 & 486.63 & 25.2M & 2113 \\
            Faster RCNN & TPAMI 2016 & Event Count & No & 26.93 & 29.68 & 27.39 & 689.68 & 41.2M & 3962 \\
            DETR & ICLR 2020 & Event Count & No & 30.35 & 33.63 & 31.64 & 631.37 & 39.8M & 3136 \\
            YOLOv10-S & NIPS 2025 & Event Count & No & 32.55 & 33.39 & 32.18 & 589.67 & 7.3M & 1627 \\ 
            \midrule
            
            EMS-YOLO & ICCV 2023 & SNN & Yes & 36.77 & 42.92 & 50.68 & 112.36 & 3.3M & 1229 \\
            Spike-YOLO & ECCV 2024 & SNN & Yes & 43.94 & 48.26 & 59.62 & 55.38 & 69.0M & 1883 \\
            GET & ICCV 2023 & Group Token & Yes & 40.31 & 48.91 & 60.73 & 46.35 & 18.4M & 2168 \\
            RED & NeurIPS 2020 & Voxel Grid & Yes & 35.99 & 45.54 & 53.76 & 102.27 & 24.1M & 3427 \\
            RVT & CVPR 2023 & Voxel Grid & Yes & 43.21 & 51.38 & 60.35 & 55.68 & 9.9M & 1737 \\
            SAST & CVPR 2024 & Voxel Grid & Yes & 34.31 & 40.22 & 51.21 & 150.32 & 18.5M & 3075 \\ 
            \midrule
            
            KPConv & ICCV 2019 & Points & Yes & 48.19 & 57.28 & 68.59 & 16.32 & 50.1M & 562 \\
            RandLA-Net & CVPR 2020 & Points & Yes & 50.32 & 59.29 & 70.56 & 6.95 & \textbf{1.2M} & 353 \\
            COSeg & CVPR 2024 & Points & Yes & 51.89 & 60.93 & 71.32 & 9.21 & 23.4M & 364 \\ 
            EV-SpSegNet & ICCV 2025 & Points & Yes & \underline{55.18} & \underline{65.02} & \underline{77.53} & \underline{1.63} & 4.0M & \textbf{36} \\
            
            % Ours 行：背景色高亮
            \rowcolor{lightgreen} 
            \textbf{Ours} & - & Points & Yes & \textbf{75.90} & \textbf{80.05} & \textbf{91.84} & \textbf{0.76} & \underline{2.9M} & \underline{58} \\ 
            \bottomrule
        \end{tabular}%
    }
    
\end{table*}

\begin{figure*}[t]
  \centering
  \includegraphics[width=0.95\textwidth]{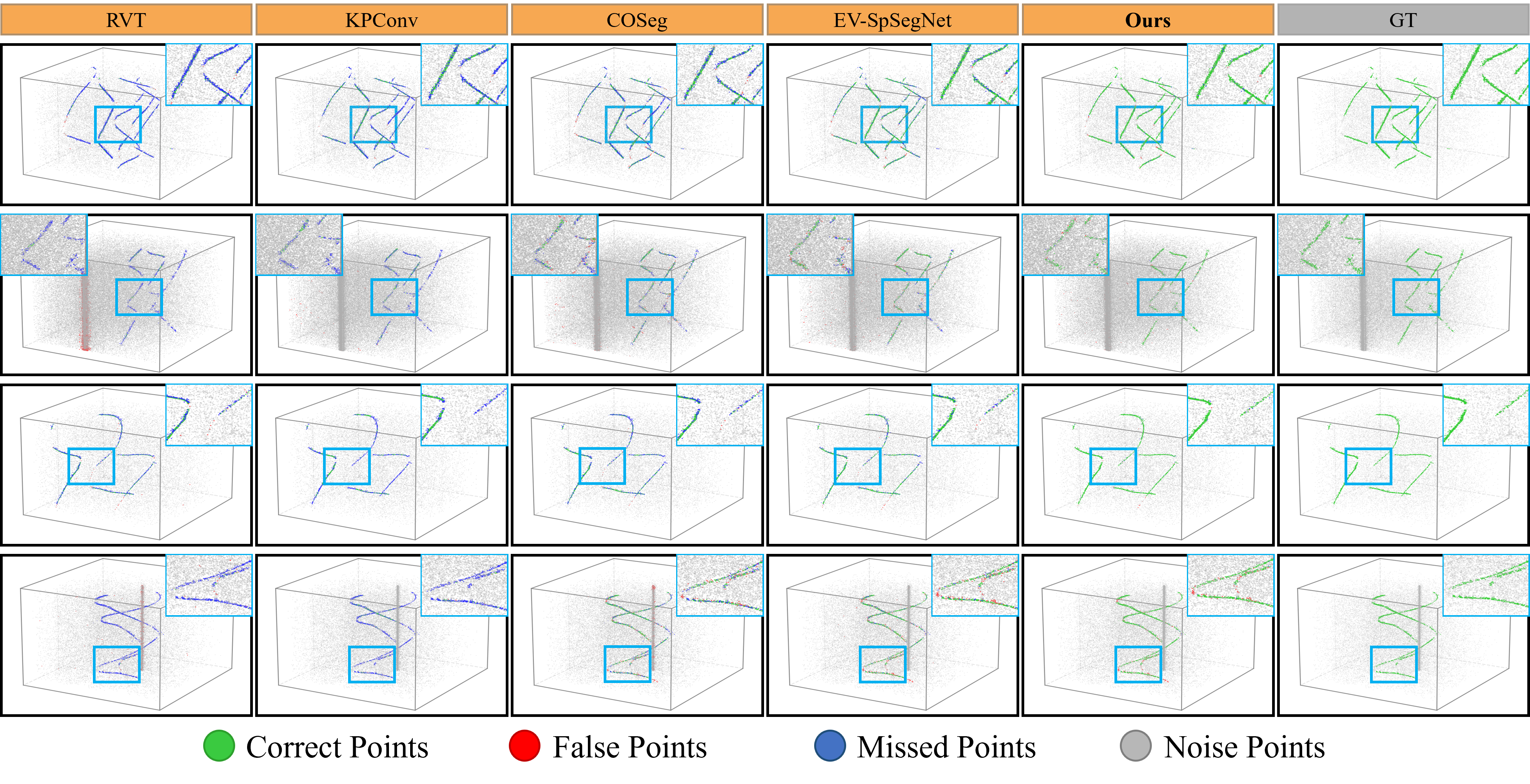}
  \caption{Qualitative results of different methods. Correct detections are marked in green, false detections in red, missed detections in blue, and noise points in gray.}
  \label{fig:expersion-vis}
\end{figure*}

We divide the compared methods into three groups and select several state-of-the-art approaches in each group. The first group consists of frame-based generic object detectors, including SSD \cite{liu2016ssd}, Faster R-CNN \cite{ren2015faster}, Deformable DETR \cite{zhu2020deformable}, and YOLOv10 \cite{wang2024yolov10}. The second group covers event-based detection methods, including RVT \cite{gehrig2023recurrent}, SAST \cite{peng2024scene}, GET \cite{peng2023get}, RED \cite{perot2020learning}, EMS-YOLO \cite{su2023deep}, and Spike-YOLO \cite{luo2024integer}. The third group contains point-cloud segmentation methods, including KPConv \cite{thomas2019kpconv}, RandLA-Net \cite{hu2020randla}, COSeg \cite{an2024rethinking}, and EV-SpSegNet \cite{chen2025event}. For the frame-based methods, the input is an event frame aggregated over a 50 ms time window.

% 如表 1 所示，各方法性能呈现明显的层级差异。基于帧的方法表现最弱，因为将 50 ms 窗口压缩为单帧会抹平窗口内的时序联系，不利于保留微弱响应之间的关联。SNN 方法虽引入时序累积，但其依赖稳定的重复激活，而间歇性且伴随漂移的微弱响应难以形成有效累积，因此整体提升有限。基于点云的方法借助稀疏几何结构显著提升了性能，但仍缺乏对微弱响应传播一致性的显式约束。相比之下，我们的方法利用物理平流一致性对特征演化施加约束，强化了符合输运规律的微弱轨迹，使 IoU 达到 75.90% 且 Pd 达到 91.84%。同时，底层稀疏卷积的实现确保了较高的推理效率。
Table~\ref{tab:comparison} reports the results of frame-based, event-based, and point-based detectors. 
Frame-based methods perform the worst, as 50 ms event aggregation removes much of the within-window temporal structure. 
Event-based models improve temporal modeling but remain limited in low-SNR scenes with heavy clutter.
Point-based segmentation baselines perform best among existing methods, with EV-SpSegNet reaching 55.18\% IoU and 77.53\% $P_d$. 
PACT further increases the IoU to 75.90\% and $P_d$ to 91.84\%, while reducing $F_a$ from $1.63$ to $0.76$ ($\times10^{-4}$). With 2.9M parameters and 58 ms inference time per window, PACT achieves this gain without increasing model complexity.

% 图 4 给出了不同方法的定性对比结果。基于帧的方法整体性能过弱，缺乏有效对比性，因此图中不再展示。作为基于事件的方法代表，RVT 在多数场景中漏检较多并伴随零散误检，说明其在强背景干扰下的鲁棒性有限。KPConv 与 COSeg 通过更好地保持稀疏几何结构降低了误检，使目标附近响应更集中，但在噪声较强的样例中仍会出现轨迹断裂与贴附背景的误检。EV SpSegNet 整体更稳定，但部分场景仍残留背景触发且连续性不足，表明仅依赖几何结构仍难以区分相似触发。相比之下，我们的方法得到更连续、更集中的正确检测点，漏检与误检更少，说明平流一致性有助于在强背景干扰下保留微弱事件响应并抑制不一致触发。
Qualitative Results: Fig.~\ref{fig:expersion-vis} shows representative qualitative results.
RVT frequently misses targets and produces scattered detections under clutter.
Point-based methods preserve sparse geometry better, but still suffer from trajectory breaks and residual background triggers.
PACT produces denser and more continuous target trajectories with fewer missed and false detections, confirming the benefit of advection-consistent propagation.

\subsection{Ablation Studies}
\begin{table}[t]
    \centering
    \caption{Ablation study on framework scheme.}
    \label{tab:ablation_study}
    \setlength{\tabcolsep}{2pt} 
    \renewcommand{\arraystretch}{1} 
    \begin{tabular}{cccccccc}
        \toprule
        T-FE & ATC & A-FR & IoU$\uparrow$ & ACC$\uparrow$ & $P_{d}\uparrow$ & $F_{a}\downarrow$ & \#Params. \\
        \midrule
        \textbf{}
         &  &  & 59.90 & 63.18 & 78.80 & 0.76 & 2.3M \\
        \midrule % 用分割线隔开不同组的实验
        
        % 单模块添加
        \checkmark &  &  & 66.54 & 69.67 & 84.48 & 0.64 & 2.6M \\
         & \checkmark &  & 55.11 & 69.95 & 83.35 & 4.76 & 2.4M \\
         &  & \checkmark & 63.26 & 65.34 & 79.97 & 0.37 & 2.5M \\
        \midrule
        
        % 组合模块
        \checkmark & \checkmark &  & 71.95 & 73.48 & 85.61 & \textbf{0.18} & 2.7M \\

        \checkmark & & \checkmark & 70.55 & 76.04 & 88.76 & 1.54 & 2.8M \\
        
        % 全部模块 (最佳结果加粗)
        \checkmark   & \checkmark & \checkmark & \textbf{75.90} & \textbf{80.05} & \textbf{91.84} & 0.76 & 2.9M \\
        \bottomrule
    \end{tabular}
\end{table}
We evaluate the physical formulation of PACT in Table~\ref{tab:ablation_study}.
The Baseline has no explicit transport constraint, so weak responses remain temporally fragile and are easily overwhelmed.
When ATC is attached to the Baseline, the false alarm rate rises to $4.76\times10^{-4}$.
This indicates that enforcing advection consistency on unstructured features induces spurious temporal correspondences, because the baseline features provide no a stable substrate for transport consistency.
With only T-FE or only A-FR, the model behaves conservatively.
These variants suppress noise effectively, yet they do not aggregate intermittent responses into coherent temporal traces, which limits the detection gain.
The full benefit emerges only after closing the physical loop.
T-FE produces motion-consistent features, and ATC imposes a transport-feasibility constraint on their propagation. 
A-FR then extends the consistent responses along the estimated flow to form stable temporal traces.
Overall, the ablation provides evidence for a coupled cycle of extraction, constraint, and propagation for sustaining weak temporal traces.

\subsection{Advection-Guided Representation Evolution}
\label{sec:exp_transport_emergence}
\begin{figure*}[t]
  \centering
  \includegraphics[width=0.95\textwidth]{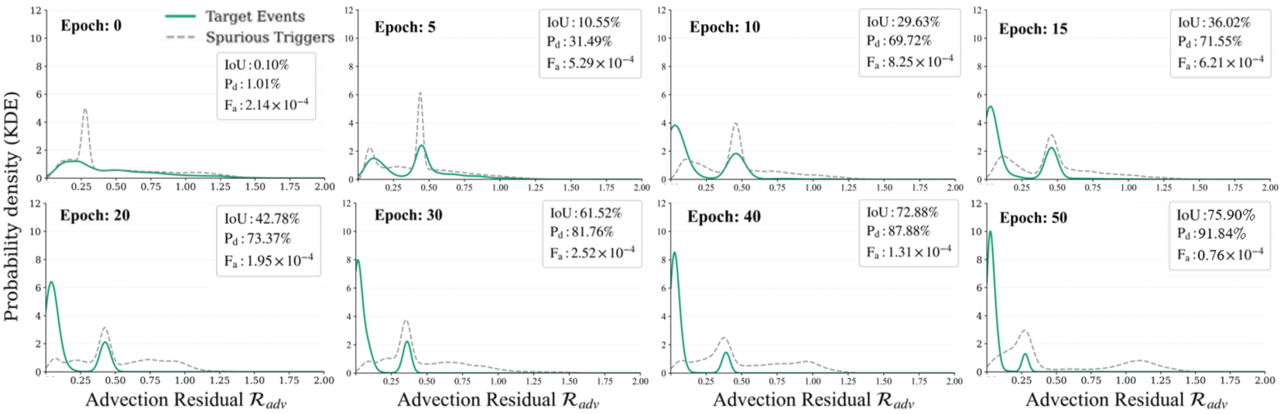}
  \caption{Advection residual density across training stages. Kernel density estimates show the advection residual $\mathcal{R}_{adv}$ for target events and spurious triggers, with the corresponding detection metrics reported next to each stage.}
  \label{fig:exp-res}
\end{figure*}
PACT uses advection consistency to preserve weak, fragmented temporal responses while suppressing triggers that do not support coherent transport. 
A key question is whether this constraint actually reshapes the internal representation, rather than merely improving the final metrics. 
To examine this, we track the distribution of advection residuals during training and test whether foreground responses progressively concentrate at low residuals while background activity remains broadly distributed at higher residuals.

For each checkpoint, we measure the advection residual between features before and after transport under the estimated velocity field,
\begin{equation}
\mathcal{R}_{adv}=\left\|\mathbf{F}-\mathcal{T}_{\mathbf{v}}(\mathbf{F})\right\|,
\end{equation}
and plot kernel density estimates for foreground (green) and background (gray) responses using ground-truth labels. 
Fig.~\ref{fig:exp-res} shows four stages from initialization to convergence, together with the corresponding detection metrics.

At early stages, foreground and background residuals largely overlap, indicating that transport does not yet provide reliable alignment for weak responses. 
As training proceeds, foreground residuals form a sharp peak near zero and separate clearly from the background, whereas background residuals remain broad at higher values. 
A near-zero residual means that the transported prediction agrees with the observation, allowing intermittent target activations to accumulate into coherent temporal traces, while clutter remains in the high-residual regime. 
Overall, the growing separation is consistent with the metric improvement and provides direct evidence that transport consistency stabilizes weak traces under heavy clutter.

% \begin{table}[h]
% \centering
% \small
% \caption{American League West Standings, 2024}
% \begin{tabular}{lccccccc}
%  Team &  W & L  & Pct & {\bf GB} & Home & Away & L10 \\
%  \hline
%  Astros & 88 & 73 & .547 & {\bf -} & 46-35 & 42-38 & 6-4 \\
%  Mariners & 85 & 77 & .525 & {\bf 3.5} & 49-32 & 36-45 & 8-2 \\
%  Rangers & 78 & 84 & .481 & {\bf 10.5} & 44-37 & 34-47 & 5-5 \\
%  Athletics & 69 & 93 & .426 & {\bf 19.5} & 38-43 & 31-50 & 3-7 \\
%  Angels & 63 & 99 & .389 & {\bf 25.5} & 32-49 & 31-50 & 1-9
% \end{tabular}
% \label{table: alweststandings2024}
% \end{table}

\subsection{Temporal Continuity Analysis}
\begin{figure}[htp]
  \centering
  \includegraphics[width=0.6\linewidth]{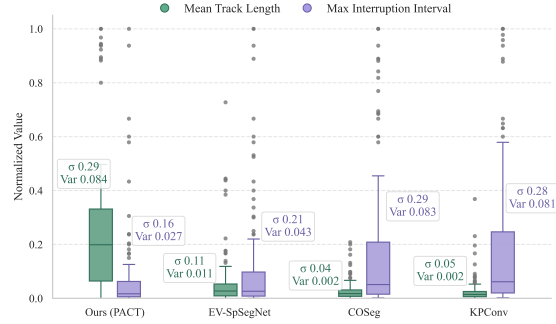}
  \caption{Comparison of temporal continuity statistics across methods. We report the average length of continuous successful detection and the maximum length of continuous detection loss per target.}
  \label{fig:temporal}
\end{figure}

\begin{figure}[t]
  \centering
  \includegraphics[width=0.95\linewidth]{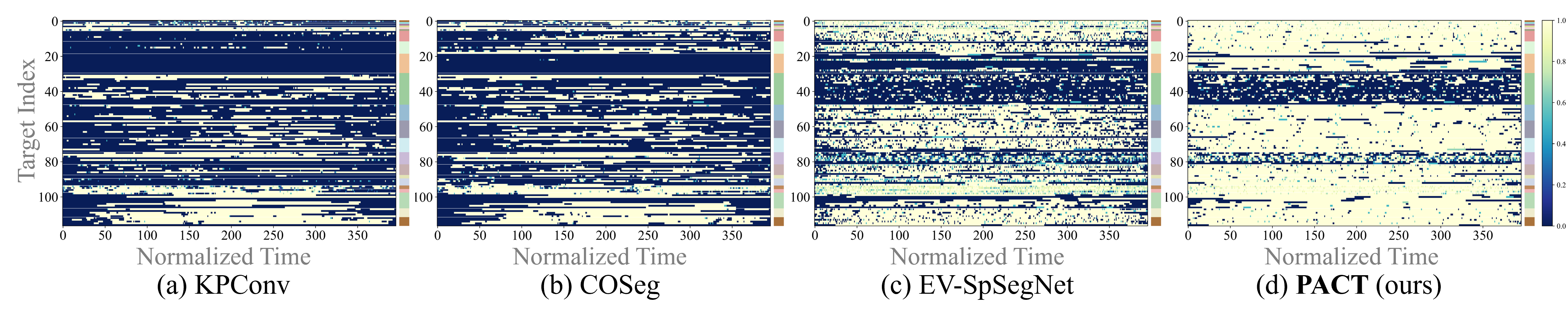}
  \caption{Comparison of temporal continuity visualizations across methods. We align each target’s detection trajectory to a unified time scale and visualize the hit ratio at each time step with color, where dark blue indicates no hit and yellow indicates a higher hit ratio. The right-side color strip groups targets by scene, with each block corresponding to one scenario, which makes it easy to compare continuous detection segments and interruption patterns across methods.}
  \label{fig:temporal_map}
\end{figure}
To quantitatively evaluate temporal consistency, we analyze the distribution of continuity metrics over all test sequences.
For each target, we convert the frame-wise hit status into a binary hit sequence and compute two complementary measures in Fig.~\ref{fig:temporal}.
The first metric is the mean track length, defined as the average length of continuous successful detection segments.
The second metric is the max interruption interval, defined as the maximum length of consecutive missed detections.
Both metrics are normalized by the ground-truth target duration so that values lie in $[0,1]$, enabling fair comparisons across targets with different temporal extents.

Fig.~\ref{fig:temporal} shows that our method yields stronger temporal continuity than the baselines.
For mean track length, our method attains a higher median and a distribution shifted toward larger values, whereas EV-SpSegNet~\cite{chen2025event}, COSeg~\cite{an2024rethinking}, and KPConv~\cite{thomas2019kpconv} concentrate at smaller values.
This indicates that baseline predictions are frequently fragmented into short segments that appear only when local evidence is strong and break once the evidence weakens.
In contrast, our approach maintains connected detections over longer spans, producing more coherent trajectories.

For max interruption interval, our method achieves a lower mean and reduced dispersion.
For example, the standard deviation decreases from $0.28$ for KPConv~\cite{thomas2019kpconv} to $0.15$ for our method.
These results align with our design motivation.
By modeling advection-based transport, our method preserves temporal connections through physics-consistent constraints, which helps bridge gaps caused by fluctuating event rates rather than relying on transient peaks.

Beyond the statistics, Fig.~\ref{fig:temporal_map} provides a complementary visual validation.
We align each target’s detection trajectory to the normalized target duration and visualize the frame-wise hit ratio, defined as the fraction of ground-truth target points correctly detected at each frame.
Colors closer to blue indicate fewer hits, while warmer colors indicate more hits.
Across targets and scenes, the baselines often exhibit bursty and intermittent responses, with short high-hit segments separated by frequent low-hit periods.
Our method instead forms longer and more continuous high-hit bands and markedly reduces prolonged low-hit stretches, suggesting that the predictions remain temporally connected rather than driven by brief local spikes.

Crucially, the $20\%$ relative IoU improvement in Table~\ref{tab:comparison} is consistent with this enhanced temporal integrity.
Longer continuous detections and fewer extended interruptions reduce temporal breakages and fragmented accumulation over time.
As a result, predicted regions evolve more coherently across the sequence and yield more complete masks even when foreground evidence is intermittent.

% 我们的方法在整体实验中显著提升了时序连续性与 IoU。我们随后在时序连续性可视化中发现一个例外现象：在 target index 约 40 对应的一组场景里，我们的方法与 EV-SpSegNet 的差距并没有明显拉开。我们因此把这一组场景单独拿出来做机制探索，目的是找出“什么时候平流输运带来的额外增益会变小”。

% 我们首先用案例可视化给出直接证据。该场景在同一帧内同时出现多个不同 ID 的目标，并且这些目标在相邻帧之间的位移幅度与方向差异很大，属于典型的多轨迹共存且速度差异显著的运动模式。基于这一观察，我们给出机制解释：我们的方法采用一阶速度场做平流输运对齐，它隐含了短时间窗内近似匀速的假设。当同一时间窗内存在多条轨迹且速度差异大，单一的一阶输运难以同时对齐所有目标，部分目标的时间聚合会被拉散或稀释，因此连续性优势在该场景组被削弱。

% 为了避免“只解释一个例子”，我们进一步做全场景统计验证。我们在每个场景内计算两类指标：目标共存重叠率，用来量化同一批帧内多目标同时出现的程度；速度离散度，用来量化共存目标之间的平均速度差异。统计结果表明，重叠率更高且速度离散度更大的场景，往往对应更小的连续性增益或更难拉开与基线的差距。这一规律与一阶输运在多轨迹异速运动下的局限性一致，从而把“现象—机制—统计”闭环起来。

% 最后，我们把这一发现转化为后续改进方向。由于问题来自缺少二阶运动建模，我们计划引入加速度场来更新速度场，或采用多假设输运机制以适配同窗内的多轨迹运动，从而在该类场景中进一步扩大连续性优势。

\begin{figure}[t]
  \centering
  \includegraphics[width=0.85\linewidth]{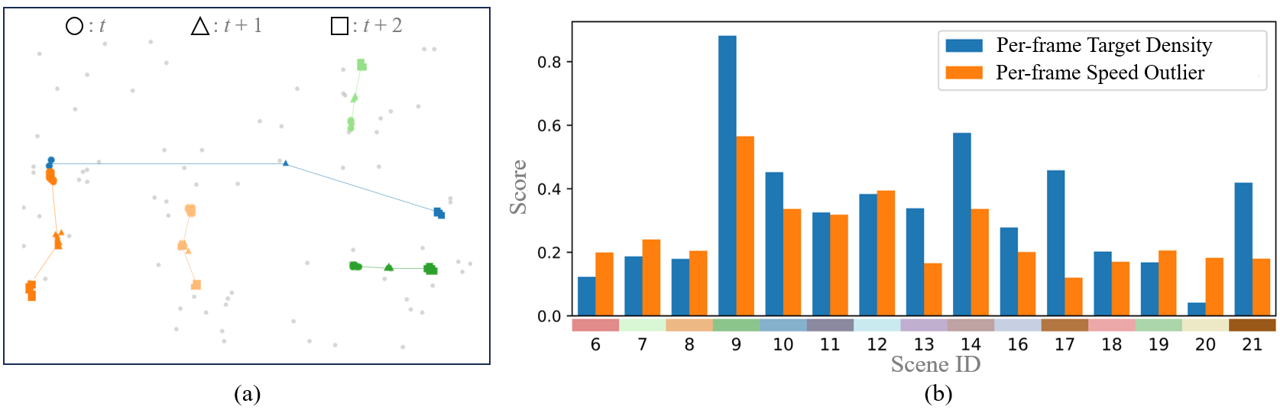}
  \caption{Mechanism study of challenging scenes.
(a) Three-frame visualization of a representative case. Marker shape indicates time (circle: $t$, triangle: $t{+}1$, square: $t{+}2$), and color indicates target ID. Co-occurring targets exhibit markedly different displacements within the same window.
(b) Scene-level statistics after removing simple scenes without temporal target co-occurrence, since they do not form competing motions within a window. Per-frame Target Density counts how many target IDs are active simultaneously in a frame. Per-frame Speed Outlier measures how strongly the fastest targets deviate from the typical motion among co-occurring targets.}
  \label{fig:multi_traj}
\end{figure}

\subsection{Multi-Target Motion with Large Velocity Variation}
\label{sec:mechanism_multitraj}

Fig.~\ref{fig:temporal_map} shows that PACT improves temporal continuity in most scenes, but the gain over EV-SpSegNet shrinks for the scene group around target index $\sim$40.
We analyze this group to identify motion patterns under which single-field advection brings limited additional benefit.

Qualitative evidence (Fig.~\ref{fig:multi_traj}(a)).
We visualize a three-frame window, where marker shapes indicate time and colors indicate target IDs.
Multiple targets co-occur within the same window, yet their inter-frame displacements differ substantially.
PACT propagates evidence with a single first-order velocity field; this works well when co-occurring targets share similar motion, but it cannot align multiple trajectories with markedly different velocities.
As a result, temporal aggregation weakens for a subset of targets, reducing the continuity gain in this group.

Scene-level statistics (Fig.~\ref{fig:multi_traj}(b)).
We compute statistics only on scenes with temporal target co-occurrence, and each color block matches a scene group in Fig.~\ref{fig:temporal_map}.
The target-index-$\sim$40 group (Scene~9) exhibits higher \emph{Per-frame Target Density} and the highest \emph{Per-frame Speed Outlierness}.
This indicates that more targets compete within the same window and that a few targets frequently move much faster than the rest, which makes a single first-order field insufficient to align all trajectories consistently.
The statistics agree with the reduced margin in Fig.~\ref{fig:temporal_map} and support the above mechanism.

Overall, the hard cases are characterized by within-window motion inconsistency, especially the presence of fast outliers or abrupt speed changes.
Extending the transport model with time-varying motion may better handle such mixed-motion windows.

\section{Conclusion and Future Work}
In this work, we formulate event streams as a continuously evolving spatiotemporal process. 
Under heavy clutter, the critical challenge is not merely sparsity, but the lack of an explicit propagation mechanism.
We cast the mechanism as a local transport process, where motion continuity is expressed as approximate feature conservation along an estimated velocity field. 
Guided by this view, PACT enforces advection consistency to sustain weak responses and suppress noise.

In our experiments, we examine the mechanism by tracking the evolution of transport-residual distributions across checkpoints. 
As the model converges, target responses increasingly concentrate in a low-residual regime, whereas background triggers remain broadly distributed at higher residuals. 
The growing separation mirrors the improvement in downstream metrics, providing indirect evidence that the model relies on transport validity to maintain weak responses instead of memorizing static patterns.

However, the current formulation models transport with locally constant velocity, which is best suited to short, near-linear motion. 
When acceleration or other non-linear dynamics become prominent, this approximation may become inaccurate. 
Future work will explore transport models beyond first-order motion so that the same consistency principle extends to more complex dynamics.

% \section*{Acknowledgements}
% Please insert your acknowledgments here.

\bibliographystyle{splncs04}
\bibliography{main}

@article{lichtsteiner2008temporal,
  title   = {A 128{\texttimes}128 120 dB 15 {\textmu}s Latency Asynchronous Temporal Contrast Vision Sensor},
  author  = {Lichtsteiner, Patrick and Posch, Christoph and Delbruck, Tobi},
  journal = {IEEE Journal of Solid-State Circuits},
  volume  = {43},
  number  = {2},
  pages   = {566--576},
  year    = {2008}
}

@article{gallego2022survey,
  title   = {Event-Based Vision: A Survey},
  author  = {Gallego, Guillermo and Delbruck, Tobi and Orchard, Garrick and Bartolozzi, Chiara and Taba, Brian and Censi, Andrea and Leutenegger, Stefan and Davison, Andrew J. and Conradt, J{\"o}rg and Daniilidis, Kostas and Scaramuzza, Davide},
  journal = {IEEE Transactions on Pattern Analysis and Machine Intelligence},
  volume  = {44},
  number  = {1},
  pages   = {154--180},
  year    = {2022}
}

@inproceedings{kim2016realtime,
  title     = {Real-Time 3D Reconstruction and 6-DoF Tracking with an Event Camera},
  author    = {Kim, Hanme and Leutenegger, Stefan and Davison, Andrew J.},
  booktitle = {Proceedings of the European Conference on Computer Vision},
  series    = {Lecture Notes in Computer Science},
  volume    = {9910},
  pages     = {349--364},
  year      = {2016},
  publisher = {Springer}
}

@article{pan2022high,
  title   = {High Frame Rate Video Reconstruction Based on an Event Camera},
  author  = {Pan, Liyuan and Hartley, Richard and Scheerlinck, Cedric and Liu, Miaomiao and Yu, Xin and Dai, Yuchao},
  journal = {IEEE Transactions on Pattern Analysis and Machine Intelligence},
  volume  = {44},
  number  = {5},
  pages   = {2519--2533},
  year    = {2022}
}

@article{shariff2024event,
  title   = {Event Cameras in Automotive Sensing: A Review},
  author  = {Shariff, Waseem and Dilmaghani, Mehdi Sefidgar and Kielty, Paul and Moustafa, Mohamed and Lemley, Joe and Corcoran, Peter},
  journal = {IEEE Access},
  volume  = {12},
  pages   = {51275--51306},
  year    = {2024}
}

@inproceedings{Maqueda2018event,
  title     = {Event-Based Vision Meets Deep Learning on Steering Prediction for Self-Driving Cars},
  author    = {Maqueda, Ana I. and Loquercio, Antonio and Gallego, Guillermo and Garc{\'\i}a, Narciso and Scaramuzza, Davide},
  booktitle = {Proceedings of the IEEE/CVF Conference on Computer Vision and Pattern Recognition},
  year      = {2018},
  pages     = {5419--5427}
}

@article{lagorce2016hots,
  title   = {HOTS: A Hierarchy of Event-Based Time-Surfaces for Pattern Recognition},
  author  = {Lagorce, Xavier and Orchard, Garrick and Galluppi, Francesco and Shi, Bertram E. and Benosman, Ryad B.},
  journal = {IEEE Transactions on Pattern Analysis and Machine Intelligence},
  volume  = {39},
  number  = {7},
  pages   = {1346--1359},
  year    = {2017}
}

@inproceedings{gehrig2019end,
  title     = {End-to-End Learning of Representations for Asynchronous Event-Based Data},
  author    = {Gehrig, Daniel and Loquercio, Antonio and Derpanis, Konstantinos G. and Scaramuzza, Davide},
  booktitle = {Proceedings of the IEEE/CVF International Conference on Computer Vision},
  pages     = {5632--5642},
  year      = {2019}
}

@article{rebecq2021high,
  title   = {High Speed and High Dynamic Range Video with an Event Camera},
  author  = {Rebecq, Henri and Ranftl, Ren{\'e} and Koltun, Vladlen and Scaramuzza, Davide},
  journal = {IEEE Transactions on Pattern Analysis and Machine Intelligence},
  volume  = {43},
  number  = {6},
  pages   = {1964--1980},
  year    = {2021}
}

@inproceedings{zhu2019unsupervised,
  title     = {Unsupervised Event-Based Learning of Optical Flow, Depth, and Egomotion},
  author    = {Zhu, Alex Zihao and Yuan, Liangzhe and Chaney, Kenneth and Daniilidis, Kostas},
  booktitle = {Proceedings of the IEEE/CVF Conference on Computer Vision and Pattern Recognition},
  pages     = {989--997},
  year      = {2019}
}

@inproceedings{perot2020learning,
  title     = {Learning to Detect Objects with a 1 Megapixel Event Camera},
  author    = {Perot, Etienne and {de Tournemire}, Pierre and Nitti, Davide and Masci, Jonathan and Sironi, Amos},
  booktitle = {Advances in Neural Information Processing Systems},
  volume    = {33},
  pages     = {16639--16652},
  year      = {2020}
}

@inproceedings{gehrig2023recurrent,
  title     = {Recurrent Vision Transformers for Object Detection with Event Cameras},
  author    = {Gehrig, Mathias and Scaramuzza, Davide},
  booktitle = {Proceedings of the IEEE/CVF Conference on Computer Vision and Pattern Recognition},
  pages     = {13884--13893},
  year      = {2023}
}

@inproceedings{su2023deep,
  title     = {Deep Directly-Trained Spiking Neural Networks for Object Detection},
  author    = {Su, Qiaoyi and Chou, Yuhong and Hu, Yifan and Li, Jianing and Mei, Shijie and Zhang, Ziyang and Li, Guoqi},
  booktitle = {Proceedings of the IEEE/CVF International Conference on Computer Vision},
  pages     = {6555--6565},
  year      = {2023}
}

@inproceedings{zhang2022spiking,
  title     = {Spiking Transformers for Event-Based Single Object Tracking},
  author    = {Zhang, Jiqing and Dong, Bo and Zhang, Haiwei and Ding, Jianchuan and Heide, Felix and Yin, Baocai and Yang, Xin},
  booktitle = {Proceedings of the IEEE/CVF Conference on Computer Vision and Pattern Recognition},
  pages     = {8801--8810},
  year      = {2022},
  publisher = {IEEE}
}

@inproceedings{schaefer2022aegnn,
  title     = {AEGNN: Asynchronous Event-Based Graph Neural Networks},
  author    = {Schaefer, Simon and Gehrig, Daniel and Scaramuzza, Davide},
  booktitle = {Proceedings of the IEEE/CVF Conference on Computer Vision and Pattern Recognition},
  pages     = {12371--12381},
  year      = {2022},
  publisher = {IEEE}
}

@inproceedings{bi2019graph,
  title     = {Graph-Based Object Classification for Neuromorphic Vision Sensing},
  author    = {Bi, Yin and Chadha, Aaron and Abbas, Alhabib and Bourtsoulatze, Eirina and Andreopoulos, Yiannis},
  booktitle = {Proceedings of the IEEE/CVF International Conference on Computer Vision},
  pages     = {491--501},
  year      = {2019}
}

@inproceedings{gallego2018unifying,
  title     = {A Unifying Contrast Maximization Framework for Event Cameras, with Applications to Motion, Depth, and Optical Flow Estimation},
  author    = {Gallego, Guillermo and Rebecq, Henri and Scaramuzza, Davide},
  booktitle = {Proceedings of the IEEE Conference on Computer Vision and Pattern Recognition},
  pages     = {3867--3876},
  year      = {2018}
}

@inproceedings{stoffregen2019event,
  title     = {Event-Based Motion Segmentation by Motion Compensation},
  author    = {Stoffregen, Timo and Gallego, Guillermo and Drummond, Tom and Kleeman, Lindsay and Scaramuzza, Davide},
  booktitle = {Proceedings of the IEEE International Conference on Computer Vision},
  pages     = {7244--7253},
  year      = {2019}
}

@inproceedings{mitrokhin2018event,
  title     = {Event-Based Moving Object Detection and Tracking},
  author    = {Mitrokhin, Anton and Ferm{\"u}ller, Cornelia and Parameshwara, Chethan and Aloimonos, Yiannis},
  booktitle = {Proceedings of the IEEE/RSJ International Conference on Intelligent Robots and Systems},
  pages     = {1--9},
  year      = {2018}
}

@inproceedings{wang2025object,
  title     = {Object Detection using Event Camera: A MoE Heat Conduction based Detector and A New Benchmark Dataset},
  author    = {Wang, Xiao and Jin, Yu and Wu, Wentao and Zhang, Wei and Zhu, Lin and Jiang, Bo and Tian, Yonghong},
  booktitle = {Proceedings of the IEEE/CVF Conference on Computer Vision and Pattern Recognition},
  pages     = {29321--29330},
  year      = {2025}
}

@inproceedings{sekikawa2023live,
  title     = {Tangentially Elongated Gaussian Belief Propagation for Event-based Incremental Optical Flow Estimation},
  author    = {Sekikawa, Yusuke and Nagata, Jun},
  booktitle = {Proceedings of the IEEE/CVF Conference on Computer Vision and Pattern Recognition Workshops},
  pages     = {21940--21949},
  year      = {2023}
}

@inproceedings{mondal2021moving,
  title     = {Moving Object Detection for Event-Based Vision Using Graph Spectral Clustering},
  author    = {Mondal, Anindya and Giraldo, Jhony H. and Bouwmans, Thierry and Chowdhury, Ananda S.},
  booktitle = {Proceedings of the IEEE/CVF International Conference on Computer Vision},
  pages     = {876--884},
  year      = {2021},
  publisher = {IEEE}
}

@inproceedings{liu2016ssd,
  title     = {SSD: Single Shot MultiBox Detector},
  author    = {Liu, Wei and Anguelov, Dragomir and Erhan, Dumitru and Szegedy, Christian and Reed, Scott and Fu, Cheng-Yang and Berg, Alexander C.},
  booktitle = {Proceedings of the European Conference on Computer Vision},
  pages     = {21--37},
  year      = {2016},
  publisher = {Springer}
}

@article{ren2015faster,
  title   = {Faster R-CNN: Towards Real-Time Object Detection with Region Proposal Networks},
  author  = {Ren, Shaoqing and He, Kaiming and Girshick, Ross and Sun, Jian},
  journal = {Advances in Neural Information Processing Systems},
  volume  = {28},
  year    = {2015}
}

@article{wang2024yolov10,
  title   = {YOLOv10: Real-Time End-to-End Object Detection},
  author  = {Wang, Ao and Chen, Hui and Liu, Lihao and Chen, Kai and Lin, Zijia and Han, Jungong},
  journal = {Advances in Neural Information Processing Systems},
  volume  = {37},
  pages   = {107984--108011},
  year    = {2024}
}

@misc{zhu2020deformable,
  title  = {Deformable DETR: Deformable Transformers for End-to-End Object Detection},
  author = {Zhu, Xizhou and Su, Weijie and Lu, Lewei and Li, Bin and Wang, Xiaogang and Dai, Jifeng},
  year   = {2020},
  note   = {arXiv:2010.04159}
}

@inproceedings{mitrokhin2020learning,
  title     = {Learning Visual Motion Segmentation Using Event Surfaces},
  author    = {Mitrokhin, Anton and Hua, Zhiyuan and Ferm{\"u}ller, Cornelia and Aloimonos, Yiannis},
  booktitle = {Proceedings of the IEEE/CVF Conference on Computer Vision and Pattern Recognition},
  pages     = {14414--14423},
  year      = {2020}
}

@inproceedings{yang2022querydet,
  title     = {QueryDet: Cascaded Sparse Query for Accelerating High-Resolution Small Object Detection},
  author    = {Yang, Chenhongyi and Huang, Zehao and Wang, Naiyan},
  booktitle = {Proceedings of the IEEE/CVF Conference on Computer Vision and Pattern Recognition},
  pages     = {13668--13677},
  year      = {2022}
}

@inproceedings{cordone2022object,
  title     = {Object Detection with Spiking Neural Networks on Automotive Event Data},
  author    = {Cordone, Lo{\"i}c and Miramond, Beno{\^\i}t and Thierion, Philippe},
  booktitle = {Proceedings of the International Joint Conference on Neural Networks},
  pages     = {1--8},
  year      = {2022},
  publisher = {IEEE}
}

@inproceedings{luo2024integer,
  title     = {Integer-Valued Training and Spike-Driven Inference Spiking Neural Network for High-Performance and Energy-Efficient Object Detection},
  author    = {Luo, Xinhao and Yao, Man and Chou, Yuhong and Xu, Bo and Li, Guoqi},
  booktitle = {Proceedings of the European Conference on Computer Vision},
  pages     = {253--272},
  year      = {2024},
  publisher = {Springer}
}

@article{neftci2019surrogate,
  title   = {Surrogate Gradient Learning in Spiking Neural Networks: Bringing the Power of Gradient-Based Optimization to Spiking Neural Networks},
  author  = {Neftci, Emre O. and Mostafa, Hesham and Zenke, Friedemann},
  journal = {IEEE Signal Processing Magazine},
  volume  = {36},
  number  = {6},
  pages   = {51--63},
  year    = {2019}
}

@article{benosman2013event,
  title   = {Event-Based Visual Flow},
  author  = {Benosman, Ryad and Clercq, Charles and Lagorce, Xavier and Ieng, Sio-Hoi and Bartolozzi, Chiara},
  journal = {IEEE Transactions on Neural Networks and Learning Systems},
  volume  = {25},
  number  = {2},
  pages   = {407--417},
  year    = {2014}
}

@misc{chen2025event,
  title  = {Event-Based Tiny Object Detection: A Benchmark Dataset and Baseline},
  author = {Chen, Nuo and Xiao, Chao and Dai, Yimian and He, Shiman and Li, Miao and An, Wei},
  year   = {2025},
  note   = {arXiv:2506.23575}
}

@misc{kingma2014adam,
  title  = {Adam: A Method for Stochastic Optimization},
  author = {Kingma, Diederik P. and Ba, Jimmy},
  year   = {2014},
  note   = {arXiv:1412.6980}
}

@inproceedings{peng2024scene,
  title     = {Scene-Adaptive Sparse Transformer for Event-Based Object Detection},
  author    = {Peng, Yansong and Li, Hebei and Zhang, Yueyi and Sun, Xiaoyan and Wu, Feng},
  booktitle = {Proceedings of the IEEE/CVF Conference on Computer Vision and Pattern Recognition},
  pages     = {16794--16804},
  year      = {2024}
}

@inproceedings{peng2023get,
  title     = {GET: Group Event Transformer for Event-Based Vision},
  author    = {Peng, Yansong and Zhang, Yueyi and Xiong, Zhiwei and Sun, Xiaoyan and Wu, Feng},
  booktitle = {Proceedings of the IEEE/CVF International Conference on Computer Vision},
  pages     = {6038--6048},
  year      = {2023}
}

@inproceedings{thomas2019kpconv,
  title     = {KPConv: Flexible and Deformable Convolution for Point Clouds},
  author    = {Thomas, Hugues and Qi, Charles R. and Deschaud, Jean-Emmanuel and Marcotegui, Beatriz and Goulette, Fran{\c{c}}ois and Guibas, Leonidas J.},
  booktitle = {Proceedings of the IEEE/CVF International Conference on Computer Vision},
  pages     = {6411--6420},
  year      = {2019}
}

@inproceedings{hu2020randla,
  title     = {RandLA-Net: Efficient Semantic Segmentation of Large-Scale Point Clouds},
  author    = {Hu, Qingyong and Yang, Bo and Xie, Linhai and Rosa, Stefano and Guo, Yulan and Wang, Zhihua and Trigoni, Niki and Markham, Andrew},
  booktitle = {Proceedings of the IEEE/CVF Conference on Computer Vision and Pattern Recognition},
  pages     = {11108--11117},
  year      = {2020}
}

@inproceedings{an2024rethinking,
  title     = {Rethinking Few-Shot 3D Point Cloud Semantic Segmentation},
  author    = {An, Zhaochong and Sun, Guolei and Liu, Yun and Liu, Fayao and Wu, Zongwei and Wang, Dan and Van~Gool, Luc and Belongie, Serge},
  booktitle = {Proceedings of the IEEE/CVF Conference on Computer Vision and Pattern Recognition},
  pages     = {3996--4006},
  year      = {2024}
}

@inproceedings{zubic2023chaos,
  title     = {From Chaos Comes Order: Ordering Event Representations for Object Recognition and Detection},
  author    = {Zubi{\'c}, Nikola and Gehrig, Daniel and Gehrig, Mathias and Scaramuzza, Davide},
  booktitle = {Proceedings of the IEEE/CVF International Conference on Computer Vision},
  pages     = {12846--12856},
  year      = {2023},
  publisher = {IEEE}
}

@inproceedings{gallego2019focus,
  title     = {Focus Is All You Need: Loss Functions for Event-Based Vision},
  author    = {Gallego, Guillermo and Gehrig, Mathias and Scaramuzza, Davide},
  booktitle = {Proceedings of the IEEE/CVF Conference on Computer Vision and Pattern Recognition},
  year      = {2019},
  publisher = {IEEE}
}

@inproceedings{messikommer2020asyncsparse,
  title     = {Event-Based Asynchronous Sparse Convolutional Networks},
  author    = {Messikommer, Nico and Gehrig, Daniel and Loquercio, Antonio and Scaramuzza, Davide},
  booktitle = {Proceedings of the European Conference on Computer Vision},
  pages     = {415--431},
  year      = {2020},
  publisher = {Springer}
}

@misc{detournemire2020automotive,
  title  = {A Large Scale Event-based Detection Dataset for Automotive},
  author = {{de Tournemire}, Pierre and Nitti, Davide and Perot, Etienne and Migliore, Davide and Sironi, Amos},
  year   = {2020},
  note   = {arXiv:2001.08499}
}

@inproceedings{zubic2024state,
  title     = {State Space Models for Event Cameras},
  author    = {Zubic, Nikola and Gehrig, Mathias and Scaramuzza, Davide},
  booktitle = {Proceedings of the IEEE/CVF Conference on Computer Vision and Pattern Recognition},
  pages     = {5819--5828},
  year      = {2024},
  publisher = {IEEE}
}

@inproceedings{li2025asynchronous,
  title     = {Asynchronous Collaborative Graph Representation for Frames and Events},
  author    = {Li, Dianze and Li, Jianing and Liu, Xu and Fan, Xiaopeng and Tian, Yonghong},
  booktitle = {Proceedings of the IEEE/CVF Conference on Computer Vision and Pattern Recognition},
  pages     = {1655--1666},
  year      = {2025},
  publisher = {IEEE}
}

@inproceedings{torbunov2025evrt,
  title     = {EvRT-DETR: Latent Space Adaptation of Image Detectors for Event-Based Vision},
  author    = {Torbunov, Dmitrii and Ren, Yihui and Ghose, Animesh and Dim, Odera and Cui, Yonggang},
  booktitle = {Proceedings of the IEEE/CVF International Conference on Computer Vision},
  pages     = {9812--9821},
  year      = {2025},
  publisher = {IEEE}
}

@inproceedings{shiba2025simultaneous,
  title     = {Simultaneous Motion and Noise Estimation with Event Cameras},
  author    = {Shiba, Shintaro and Aoki, Yoshimitsu and Gallego, Guillermo},
  booktitle = {Proceedings of the IEEE/CVF International Conference on Computer Vision},
  pages     = {6959--6969},
  year      = {2025},
  publisher = {IEEE}
}

@inproceedings{yamaki2025iterative,
  title     = {Iterative Event-Based Motion Segmentation by Variational Contrast Maximization},
  author    = {Yamaki, Ryo and Shiba, Shintaro and Gallego, Guillermo and Aoki, Yoshimitsu},
  booktitle = {Proceedings of the IEEE/CVF Conference on Computer Vision and Pattern Recognition Workshops},
  pages     = {4918--4927},
  year      = {2025},
  publisher = {IEEE}
}

\end{document}

\title{Following the Flow: Advection-Consistent Modeling for Event-based Small Object Detection\\[0.5ex]
Supplementary Material}

\titlerunning{Supplementary Material}
\author{Anonymous Authors}
\authorrunning{Anonymous Authors}
\institute{ECCV 2026 Submission ID 13150}

\maketitle

\section{Additional Related Work}

Event-based learning has gradually moved beyond using asynchronous streams only as regular tensors or recurrent states for downstream models \cite{gehrig2019end,gehrig2023recurrent}.
More recent studies begin to examine the representation itself.
Zubi\'c \etal show that dense event representations can be explicitly ranked with Gromov-Wasserstein discrepancy for downstream tasks \cite{zubic2023chaos}, and later introduce state-space models for event cameras to replace conventional recurrence with learned temporal dynamics \cite{zubic2024state}. 
Asynchronous sparse processing has also continued to develop.
Schaefer \etal update only the affected part of an evolving spatio-temporal graph in AEGNN \cite{schaefer2022aegnn}, Messikommer \etal convert synchronous event models into asynchronous sparse ones while preserving identical outputs \cite{messikommer2020asyncsparse}, and Li \etal further couple frames and events in a unified asynchronous graph representation \cite{li2025asynchronous}. 
On the detector side, Torbunov \etal adapt image detectors to event-based vision in latent space rather than redesigning the detector from scratch \cite{torbunov2025evrt}. 
Across these developments, the emphasis remains on how event streams are represented, updated, or transferred into downstream models. 
The formulation adopted here enters at a later stage, when weak encoded responses already exist and must remain coherent over time.

Motion-guided event modeling has likewise developed beyond early event warping. 
Gallego \etal formulate event alignment through contrast maximization under a motion model, while Benosman \etal relate this motion-compensation view to event-based visual flow estimation \cite{gallego2018unifying,benosman2013event}. 
Gallego \etal further compare a broad family of focus-based losses and analyze how different objectives affect motion-related estimation \cite{gallego2019focus}. 
Motion compensation has also been used more directly for motion segmentation and moving-object analysis \cite{stoffregen2019event,mitrokhin2018event}.
More recent work begins to model motion together with degradation in the data itself: Shiba \etal estimate motion and noise jointly \cite{shiba2025simultaneous}, while Yamaki \etal extend contrast-maximization-based motion segmentation to an iterative variational framework \cite{yamaki2025iterative}. 
Within this line, motion is still used mainly to align raw events, refine compensation, or define optimization targets. 
The advection-based formulation adopted here instead uses a local velocity field after encoding, where latent features are transported and weak responses are judged by whether they remain consistent under that transport.

\begin{table*}[t]
\centering
\caption{Detailed training settings of the compared methods on EV-UAV.
% The compared methods follow the configurations reported in their original papers, as our training configuration did not yield better performance for these baselines.
Baseline methods follow the configurations reported in their original papers, as alternative training configurations did not improve their performance.}
\label{tab:additional_settings}
\small
\renewcommand{\arraystretch}{1.15}
\setlength{\tabcolsep}{3pt}
\begin{tabular}{lcccccc}
\toprule
Method & Input & Optimizer & LR & Scheduler & Batch & Epochs \\
\midrule

\multicolumn{7}{l}{\textcolor{groupfg}{\textit{Frame-based methods}}} \\
SSD & \multirow{4}{*}{50ms/frame} & SGD & 1e-3 & Step & 32 & 100 \\
Faster R-CNN &   & SGD & 1e-3 & Step & 8  & 100 \\
DETR         &   & AdamW  & 1e-4   & Step & 8  & 150 \\
YOLOv10-S    &      & SGD & 1e-2 & Linear & 16 & 100 \\
\midrule

\multicolumn{7}{l}{\textcolor{groupfg}{\textit{SNN-based methods}}} \\
EMS-YOLO     & \multirow{2}{*}{2.5s/5 slice}  & SGD & 1e-2 & Cosine & 32 & 100 \\
SpikeYOLO    &   & SGD & 1e-2 & Cosine & 20 & 50 \\
GET          & 50ms/12 groups & Adam & 2e-4 & OneCycle & 8 & 50 \\
\midrule

\multicolumn{7}{l}{\textcolor{groupfg}{\textit{Voxel-based methods}}} \\
RED          & \multirow{3}{*}{50ms/10 slice volume} & Adam   & 2e-4 & Exponential & 8 & 20 \\
RVT          &  & Adam   & 2e-4 & OneCycle & 8 & 50 \\
SAST         &   & AdamW & 5e-4 & Cosine & 16 & 50 \\
\midrule

\multicolumn{7}{l}{\textcolor{groupfg}{\textit{Point-based methods}}} \\
KPConv       & \multirow{4}{*}{8s/point stream} & Momentum & 1e-2 & Step & 10 & 400 \\
RandLA-Net   &   & Adam   & 1e-2 & Exponential & 8  & 100 \\
COSeg        &   & AdamW  & 5e-5 & Cosine & 4 & 100 \\
EV-SpSegNet  &   & Adam   & 1e-2 & Linear & 8   & 50 \\
\midrule

\multicolumn{7}{l}{\textcolor{groupfg}{\textit{Ours}}} \\
PACT (ours)  & 8s/point stream & Adam & 1e-3 & StepLR & 4 & 50 \\
\bottomrule
\end{tabular}
\end{table*}

Benchmark development shows a similar difference in emphasis. 
Perot \etal introduce the 1 Mpx dataset for general event-based detection in high-resolution automotive scenes \cite{perot2020learning}.
The earlier large-scale automotive benchmark of de Tournemire \etal provides longer driving recordings but excludes targets whose bounding-box diagonal is smaller than 30 pixels \cite{detournemire2020automotive}. 
More recently, Chen \etal shift the focus to event-based tiny object detection with EV-UAV, emphasizing extremely small targets in complex anti-UAV scenarios \cite{chen2025event}. 
This setting is closer to the one considered here, where the challenge is not only whether a target can be detected, but whether weak and fragmented evidence can remain temporally continuous once event support becomes sparse and easily interrupted.

\section{Additional Experimental Settings}
For reproducibility, we further summarize the experimental settings of the compared methods on EV-UAV. Since these baselines follow different input paradigms, we configure their inputs accordingly. The resulting input formats and training configurations are reported in Table~\ref{tab:additional_settings}.

\section{Ablation on the Number of Hypotheses $K$}
We study the effect of the hypothesis number $K$ used in the A-FR module. 
This parameter determines how many perturbed transport candidates are generated around the estimated local velocity field during advection-guided propagation. 
When $K$ is too small, the candidate set cannot adequately cover motion uncertainty, making the decoder less stable when velocity estimation is unreliable in sparsely activated regions. 
As $K$ increases, the model can evaluate a broader set of transport hypotheses and better recover fragmented responses along plausible trajectory directions. 
However, an overly large $K$ also introduces redundant candidates, which increases the risk of aggregating noisy or weakly related responses and reduces the selectivity of consistency-based weighting.

Table~\ref{tab:ablation_k} shows that the performance improves as $K$ increases from a small value, and then saturates or slightly drops when too many hypotheses are introduced. 
This suggests that $K$ should be large enough to cover motion uncertainty, but not so large that redundant candidates interfere with stable aggregation. 
Accordingly, we set $K=5$ in the final model.

\begin{table}[t]
\centering
\caption{Ablation on the number of transport hypotheses $K$.}
\label{tab:ablation_k}
\small
\setlength{\tabcolsep}{6pt}
\renewcommand{\arraystretch}{1.1}
\begin{tabular}{ccccc}
\toprule
$K$ & IoU (\%) $\uparrow$ & ACC (\%) $\uparrow$ & $P_d$ (\%) $\uparrow$ & $F_a$ ($10^{-4}$) $\downarrow$ \\
\midrule
1 & 72.23 & 74.62 & 87.80 & 1.32 \\
3 & 74.42 & 79.83 & 91.64 & 1.13 \\
5 (ours) & \textbf{75.90} & \textbf{80.05} & \textbf{91.84} & \textbf{0.76} \\
7 & 74.98 & 77.76 & 89.30 & 0.93 \\
\bottomrule
\end{tabular}
\end{table}

\section{Ablation on the Maximum Displacement Bound $\mathrm{v}_{max}$}
We further study the effect of the maximum displacement bound $\mathrm{v}_{max}$ in trajectory-consistency estimation. 
This parameter sets the upper bound of the predicted spatial displacement at each scale and determines the search range of advection-guided transport. 
If $\mathrm{v}_{max}$ is too small, the model cannot cover sufficiently large motion variations, making it difficult to connect weak yet trajectory-consistent responses across neighboring time steps. 
As a result, the reconstructed trajectories remain incomplete and fragmented. 
If $\mathrm{v}_{max}$ is too large, the transport range becomes insufficiently constrained, increasing the risk of linking unrelated activations under cluttered event patterns.

As reported in Table~\ref{tab:ablation_vmax}, the best performance is achieved with a moderate displacement bound. 
This indicates that $\mathrm{v}_{max}$ should be large enough to cover target motion, but not so large that incorrect long-range transport is introduced. 
Accordingly, we set $\mathrm{v}_{max}=5$ in the final model.

\begin{table}[t]
\centering
\caption{Ablation on the maximum displacement bound $v_{max}$.}
\label{tab:ablation_vmax}
\small
\setlength{\tabcolsep}{6pt}
\renewcommand{\arraystretch}{1.1}
\begin{tabular}{ccccc}
\toprule
$\mathrm{v}_{max}$ & IoU (\%) $\uparrow$ & ACC (\%) $\uparrow$ & $P_d$ (\%) $\uparrow$ & $F_a$ ($10^{-4}$) $\downarrow$ \\
\midrule
2.5 & 69.22 & 76.42 & 89.17 & 2.06 \\
5.0 (ours) & \textbf{75.90} & \textbf{80.05} & \textbf{91.84} & \textbf{0.76} \\
7.5 & 72.11 & 75.26 & 88.11 & 1.33 \\
10.0 & 72.38 & 74.09 & 86.85 & 1.45 \\
\bottomrule
\end{tabular}
\end{table}

\section{Robustness under Extreme Temporal Sparsity}

Event streams are inherently continuous in time, whereas most learning-based models are trained and evaluated under relatively dense event observations. 
This raises a practical concern: whether the model relies on the statistical redundancy of frequent events, or whether it exploits a more intrinsic structure that remains valid when temporal evidence becomes sparse.

In our case, the core assumption behind PACT is that feature evolution follows an advection process driven by a learned velocity field. 
If this assumption holds, feature consistency should be preserved along the predicted motion trajectories, even when intermediate events are missing. 
On the contrary, a model that mainly depends on repeated activations or dense temporal patterns would be expected to degrade rapidly once such redundancy is removed.

To probe this behavior, we perform a controlled event dropping experiment at inference time. 
Starting from the original event stream, we randomly discard $30\%$, $50\%$, and $70\%$ of events, while keeping the trained model unchanged. 
Rather than focusing solely on detection accuracy, we examine the underlying mechanism by analyzing the advection residuals before and after feature transport.

\begin{figure}[t]
  \centering
  \includegraphics[width=\linewidth]{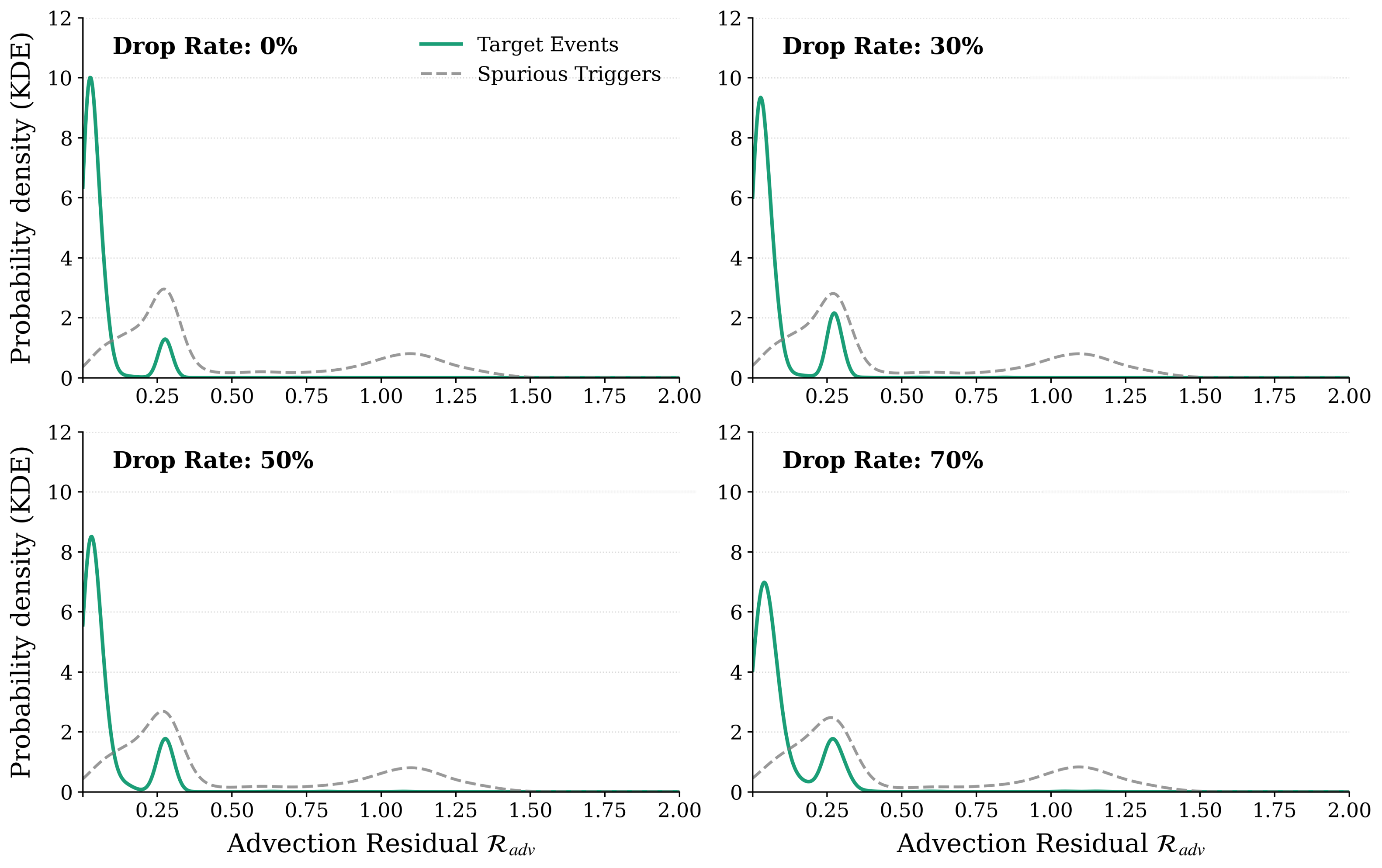}
  \caption{\textbf{Advection residual distributions under stochastic event dropping (0\%--70\%).}
  As sparsity increases, target events consistently maintain a sharp peak near zero, indicating that features are conserved along the predicted velocity field. 
  In contrast, spurious triggers exhibit much flatter and more dispersed residual distributions. 
  This suggests that PACT maintains temporal continuity through motion-consistent transport rather than relying on dense signal patterns.}
  \label{fig:robustness_kde}
\end{figure}

Specifically, we compute the kernel density estimates of the advection residual $\mathcal{R}_{adv}$ and separate the statistics according to ground-truth labels, resulting in two distributions: one corresponding to target events and the other to spurious triggers.
These distributions are visualized in Fig.~\ref{fig:robustness_kde}.

Across all drop rates, the residual distribution of target events consistently exhibits a pronounced peak near zero.
This indicates that, along the predicted velocity field, the transported features remain largely unchanged, despite substantial temporal gaps in the input stream.
Although the peak becomes slightly broader as the drop rate increases, its location remains stable even when up to $70\%$ of events are removed.
This behavior suggests that the model maintains feature continuity through motion-guided transport, rather than relying on dense temporal sampling.

In contrast, spurious triggers produce a much flatter and more dispersed residual distribution.
Since these events do not follow a coherent motion pattern, their features fail to satisfy the advection constraint, resulting in consistently higher residuals.
Importantly, the separation between the near-zero target peak and the noise distribution persists across all sparsity levels.

Taken together, these observations indicate that the robustness of PACT under extreme event dropping arises from the validity of its motion-consistent transport mechanism.
The model preserves feature coherence by propagating information along learned trajectories, instead of depending on repetitive or densely observed event patterns.

\bibliographystyle{splncs04}
\bibliography{main}